%% file: main.tex
\documentclass{article}
\usepackage{iclr2027_conference,times}
\iclrfinalcopy
\input{preamble}
\title{Beckmann World Models:\\Direct Terminal-Map Learning for\\One-Step Action-Conditioned Prediction}
\author{Adam Lee\\UC Berkeley\And Shobhit Agarwal\\Stanford University\And June Pyo Suh\\Independent Researcher}
\begin{document}
\maketitle
\lhead{Preprint}
\raggedbottom
\input{sections/abstract}
\input{sections/introduction}
\input{sections/related}
\input{sections/method}
\input{sections/experiments}
\input{sections/discussion}
\label{main-text-end}
\clearpage
\input{sections/statements}
\bibliographystyle{iclr2027_conference}
\bibliography{references,additional_references,new_references}
\clearpage
\appendix
\input{appendices/implementation}
\input{appendices/evaluation}

\input{appendices/native}
\input{appendices/diagnostics}

\input{appendices/fresh}

\end{document}

%% file: preamble.tex
\usepackage[T1]{fontenc}
\usepackage[utf8]{inputenc}
\usepackage{amsmath,amssymb,amsthm,mathtools}
\usepackage{graphicx,booktabs,tabularx,array,multirow}
\usepackage{algorithm,algpseudocode}
\usepackage{microtype,xcolor}\setcitestyle{citesep={,},yysep={,}}
\usepackage{placeins}
\usepackage[hidelinks]{hyperref}
\hypersetup{pdftitle={Beckmann World Models: Direct Terminal-Map Learning for One-Step Action-Conditioned Prediction},pdfauthor={Adam Lee, Shobhit Agarwal, JP Suh}}

\newcommand{\E}{\mathbb E}
\newcommand{\sg}{\operatorname{sg}}
\newcommand{\normD}[1]{\lVert #1\rVert_D^2}
\newcommand{\Lbtm}{\mathcal L_{\mathrm{BTM}}}
\newcommand{\Lsrc}{\mathcal L_{\mathrm{src}}}
\newcommand{\VarD}{\operatorname{Var}_D}
\newcolumntype{Y}{>{\raggedright\arraybackslash}X}

%% file: sections/abstract.tex
\begin{abstract}
Action-conditioned world models support planning and policy evaluation by predicting future states. However, learning an accurate one-step predictor can require substantial training computation. We introduce Beckmann World Models (BWM), which learn a direct map from noise, visual history, and actions to future frames without a pretrained generative teacher. We adapt the terminal-map conservation law of Beckmann Transport Models to action-conditioned video prediction, training directly from noise–data pairs without conditional populations of generated samples required by DriftWorld. However, transport consistency on interpolated inputs does not ensure accurate prediction from the pure noise used at inference. We address this gap with endpoint-aligned training that directly supervises the source prediction, emphasizes motion and spatial structure, and teaches recovery from generated histories. All components train the same map and retain one network call per future chunk. On PushT and Robomimic Can, the final model reduces 64-frame prediction MSE by $68\%$ and $57\%$, respectively, relative to same-backbone regression baseline. For policy evaluation, BWM achieves the lowest mean IoU error among the evaluated models, $0.0100$, at $24.8$\,ms per four-frame PushT prediction. On Robomimic Can, the BWM stage trained with motion supervision and generated histories achieves $3.3\times$ higher throughput and $78\%$ lower peak memory than local one-step DriftWorld. In a latent-space extension to RT-1 and Bridge-V2, equal-update controls show that feature supervision under generated histories improves perceptual prediction. These findings show how terminal-map learning can combine efficient training with accurate one-step visual prediction.
\end{abstract}

%% file: sections/introduction.tex
\section{Introduction}

An action-conditioned visual world model predicts future states given past observations and a sequence of actions. Such predictions support visual planning, policy evaluation, and learning from imagined interaction \citep{finn2016video,ebert2018visual,ha2018worldmodels,hafner2020dreamer,hafner2023dreamerv3}. Recent video simulators extend these capabilities to richer environments and manipulation tasks \citep{yang2024unisim,bruce2024genie,zhu2025irasim,zhu2025uwm,guo2026ctrlworld}. Their practical value depends on both accuracy and inference cost. 
% Evaluating many candidate actions requires many predicted futures, and a plausible video is insufficient if it places the manipulated object in the wrong configuration.

Diffusion and flow models offer expressive conditional generators, but standard samplers repeatedly evaluate a neural network to predict each future state \citep{ho2020ddpm,song2021sde,lipman2023flow,ho2022video,voleti2022mcvd}. Their sampling cost motivates learning a direct map from noise to future frames with one-step prediction. Distillation obtains such predictors from pretrained generators \citep{salimans2022progressive,yin2024dmd,yin2024dmd2}, while direct training learns them from data \citep{song2023consistency,song2024improvedcm,geng2025meanflow,geng2026imf}. Drifting models provide another route to one-step generation \citep{deng2026drifting}. DriftWorld applies drifting to action-conditioned video prediction \citep{lu2026driftworld}. These methods enable one-step prediction, but their training objectives can have substantially different computational costs. We study terminal-map learning as a way to train accurate action-conditioned predictors efficiently. 
% Our goal is to learn the conditional transformation from transitions while preserving accurate action consequences and reducing training cost.

\begin{figure}[t]
\centering
\includegraphics[width=\linewidth]{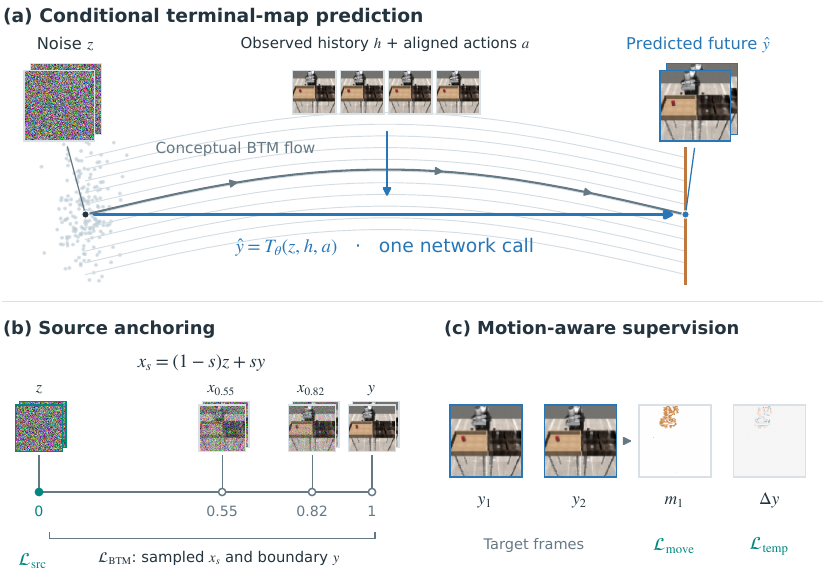}
\caption{\textbf{Conditional terminal-map prediction and training.} (a) Observed history and aligned actions condition a one-call prediction. The endpoint image shows a representative prediction on Robomimic Can. (b) Pair-based training uses $x_s=(1-s)z+sy$ and the target boundary $y$. Source anchoring supervises $T_\theta(z,h,a)$ at $s=0$. (c) Target frames define motion masks and within-chunk differences. Prediction quality is evaluated in Section~\ref{sec:experiments}.}
\label{fig:method-overview}
\end{figure}

We introduce \textbf{Beckmann World Models (BWM)}, which adapt the terminal-map formulation of Beckmann Transport Models \citep{lee2026btm} to action-conditioned visual prediction (Figure~\ref{fig:method-overview}). For fixed history and actions, a terminal map assigns each noise input the endpoint of an autonomous transport. The terminal map is constant along each autonomous transport trajectory. This conservation property yields a training objective with a stop-gradient target constructed from pairs of noise samples and ground-truth future frames. The terminal map is conditioned on observation history and actions, without an explicit transport-time input. This construction requires neither a pretrained generative teacher nor a population of generated futures for each condition. Unlike feature-space drifting, the terminal-map objective does not require a pretrained similarity encoder.

Turning this construction into an accurate world model requires supervising the inputs used at inference. The transport objective constrains the map on interpolated inputs, while boundary reconstruction teaches it to reproduce the ground-truth future frames. Neither directly supervises the prediction obtained from pure noise. In Robomimic Can, a stabilized map has $27\times$ higher prediction MSE at the source than at the target boundary. Thus, accurate boundary reconstruction can coexist with an inaccurate prediction from initial noise.

We address this gap with \emph{endpoint-aligned Beckmann training}, which directly supervises predictions from initial noise. Because much of a given scene is static, uniform pixel error can underweight the changes that distinguish action outcomes. We therefore emphasize moving pixels and temporal differences, then supervise spatial structure with frozen visual features. To prepare the map for autoregressive use, we also train subsequent predictions under detached generated histories. Together, these objectives address the discrepancy between transport training and inference under source noise and self-generated histories, without increasing inference cost. All components train the same map and preserve one forward pass per future chunk.

On PushT and Robomimic Can, the final staged model reduces 64-frame MSE by $68\%$ and $57\%$ relative to same-backbone regression. Improvements are also seen in moving regions and perceptual metrics, showing that the gains extend beyond static background reconstruction. The displayed qualitative results illustrate these gains in object pose and robot configuration (Figure~\ref{fig:rgb}).

The downstream benefit is clearest in policy evaluation. BWM estimates simulator policy scores with mean IoU error $0.0100$, compared with $0.0306$ for local DriftWorld and $0.0203$ for its released checkpoint, while predicting four frames in $24.8$\,ms. 

Compared with DriftWorld, which also uses single-pass inference, BWM substantially reduces training-step cost. On Robomimic Can, BWM achieves lower perceptual error in less cumulative training time than DriftWorld (Figure~\ref{fig:efficiency}). Complete-update profiling gives $3.3\times$ higher throughput and $78\%$ lower peak allocated memory with equal batches of windows. These measurements cover the stage trained with motion supervision and generated-history recovery, including generated prefixes, and precede the final stage with VGG feature supervision.

We further evaluate generated-history feature supervision on RT-1 and Bridge-V2 in the latent space. The hybrid objective combines terminal-map training with feature drifting. Adding feature supervision to predictions conditioned on generated histories reduces LPIPS by $34\%$ and $24\%$ against equal-update continuations from the same initial checkpoints. This extension retains one forward pass per frame and uses generated samples during training.

Our contributions are summarized below.
\begin{itemize}
\item \textbf{Conditional terminal-map learning.} We adapt Beckmann training to action-conditioned prediction and reduce the gap between target-boundary reconstruction and prediction from initial noise.

\item \textbf{Endpoint-aligned supervision.} We integrate source supervision with motion, spatial, and generated-history signals to improve one-step prediction without additional network evaluations.

\item \textbf{Training efficiency and downstream evaluation.} We evaluate complete training recipes, demonstrate improved rollouts and policy-score estimation, and test generated-history feature supervision through controlled experiments on real videos.
\end{itemize}

% The project page is available at \url{https://abecid.com/beckmann-world-models/}.

%% file: sections/related.tex
\section{Related Work}\label{sec:related}

\paragraph{Visual prediction for decision making.}
Visual world models predict future observations for candidate action sequences, enabling model-based planning \citep{finn2016video,ebert2018visual}. 
Latent world models learn compact dynamics where a policy can be trained through imagined interactions \citep{ha2018worldmodels,hafner2020dreamer,hafner2023dreamerv3}. Explicit visual prediction instead allows existing visual policies and outcome readers to interact with predicted observations. In this setting, small errors in object configuration can change the apparent consequence of an action. DIAMOND demonstrates the importance of such details for decision making and studies how diffusion sampling affects autoregressive stability \citep{alonso2024diamond}. BWM follows this explicit prediction setting and studies how to train accurate predictions while retaining a single model evaluation per chunk.

\paragraph{Controllable video simulation.}
Expressive video generation provides a foundation for visual simulation \citep{ho2022video,voleti2022mcvd,blattmann2023lvdm}, but useful robot prediction also requires a close correspondence between controls and visual outcomes. UniSim brings heterogeneous interaction data into a common action-conditioned generation framework \citep{yang2024unisim}. IRASim develops fine-grained robot action conditioning \citep{zhu2025irasim}, while Ctrl-World addresses partial observability and sustained interaction through joint camera prediction, frame-level pose conditioning, and retrieval from visual history \citep{guo2026ctrlworld}. These methods improve action conditioning, history conditioning, and consistency across camera views. BWM instead investigates how one-step generative training objectives affect prediction accuracy and training cost for single-view robotic dynamics. Our evaluations concern prediction, training cost, and policy-score estimation within the reported benchmarks.

\paragraph{Learning predictors with few sampling steps.}
Diffusion and Flow Matching learn local generative dynamics that are integrated during sampling \citep{ho2020ddpm,lipman2023flow}. Improved samplers and rectified flow reduce the cost of this integration \citep{song2021ddim,karras2022edm,liu2023rectified}. Distillation moves more computation into training by transferring a pretrained generative process to a faster student. Step-distillation, distribution matching, and adversarial supervision provide different ways to conduct that transfer \citep{salimans2022progressive,yin2024dmd,yin2024dmd2,sauer2024add}. For robot video, concurrent DIDO combines distribution matching distillation with supervision of gripper and object trajectories and DINO feature alignment before coupling the predictor to an action expert \citep{lyu2026dido}. This highlights the need to preserve interaction details as sampling is accelerated. BWM trains the predictor without a pretrained generative teacher or additional object annotations, using frozen visual features for spatial supervision. 

\paragraph{Direct map learning.}
% One-step methods differ in the quantity they learn and the relation used to supervise it. 
Consistency Models learn a common endpoint for inputs along a probability-flow trajectory \citep{song2023consistency,song2024improvedcm}. Finite-time maps and transitions across step sizes provide related ways to represent a completed portion of a flow \citep{boffi2025flowmap,frans2025shortcut}. MeanFlow instead models average velocity and derives a training identity relating it to instantaneous velocity \citep{geng2025meanflow}. BTM supplies the principle used here. Its autonomous transport defines a terminal map whose value is constant along each trajectory, motivating a stationary conservation objective without an explicit transport-time input \citep{lee2026btm}. BWM instantiates that map for conditional visual dynamics and addresses the accuracy of its prediction from initial noise. We inherit BTM's transport formulation and stop-gradient training objective. 

\paragraph{Conditional distribution updates.}
Drifting learns a direct generator by evolving its output distribution during optimization. A kernelized field attracts generated samples toward data and repels them from the current generated population \citep{deng2026drifting}. DriftWorld applies this construction separately to each visual history and action sequence, using a ground-truth future chunk and a population of generated futures \citep{lu2026driftworld}. It shares BWM's one-step interface and several endpoint regularizers, making it a close comparison for the training objective. BWM instead constructs its stop-gradient training target from noise–data pairs and a Jacobian-vector product, avoiding DriftWorld’s per-condition populations of generated samples during training. This changes the computation needed for training, although derivatives, boundary evaluations, and generated histories still incur cost. We therefore compare complete training steps and prediction quality over cumulative training time. The latent-video extension retains feature drifting and examines its combination with terminal-map training.

\paragraph{Supervision under generated histories.}
A predictor trained on ground-truth context encounters different inputs when its own outputs become the context. Self Forcing addresses this mismatch by generating autoregressive contexts during training and applying a distributional loss to the resulting video \citep{huang2025selfforcing}. DriftWorld also trains with detached generated histories \citep{lu2026driftworld}. BWM uses a detached prefix and supervises the subsequent prediction against the corresponding ground-truth future frames. This teaches recovery from context errors, without supplying physical labels for counterfactual states. Motion weighting and frozen feature losses also build on existing world-model and perceptual supervision \citep{lu2026driftworld,zhang2018lpips}. We study their integration with the terminal map and test the addition of feature supervision at generated-history endpoints through controlled continuations. Source anchoring addresses a separate mismatch at the noise input. Its squared-error supervision can favor concentrated predictions when futures are ambiguous, so paired accuracy alone does not establish calibrated conditional diversity \citep{alonso2024diamond}.

\paragraph{Evaluating world-model utility.}
Different downstream tests assess different properties of a predictor. GPC uses a world model to rank or refine action proposals from a frozen generative policy \citep{qi2026gpc}. WorldGym evaluates policies through imagined rollouts, while Ctrl-World distinguishes instruction following from successful low-level execution \citep{quevedo2025worldgym,guo2026ctrlworld}. We accordingly measure agreement with simulator policy scores separately from executed planning performance. AVDC addresses another setting, generating task-conditioned video and recovering actions through dense correspondences without action labels \citep{ko2024avdc}. Our AVDC-derived control follows the action-conditioned adaptation used in GPC and tests that video diffusion recipe within our protocol. 

%% file: sections/method.tex
\section{Beckmann World Models}\label{sec:method}

\subsection{Problem Formulation}
Let $h$ denote four observed frames, $a$ the aligned controls, and $y$ the ground-truth future chunk. We write $c=(h,a)$ and predict
\begin{equation}
\widehat y=T_\theta(z,c)=z+U_\theta(z,c),\qquad z\sim\mathcal N(0,1.35^2I).\label{eq:inference}
\end{equation}
For pixel-space prediction, we use an action-conditioned residual U-Net.
We study direct pixel-space prediction on $96\times96$ PushT and Robomimic Can videos, and latent-space prediction using an SD3 VAE on $256\times256$ RT-1 and Bridge-V2 videos. We concatenate the observed frames with the interpolated future input and condition the residual blocks on frame-aligned actions. No interpolation-time variable is supplied to the network. For a future tensor with $D$ scalar coordinates, write $\normD{v}=D^{-1}\|v\|_2^2$.

For a fixed condition, the BTM construction associates an autonomous field $b_c$ with a terminal map $T_c$. Under the support and regularity assumptions of \citet{lee2026btm}, the endpoint obeys
\begin{equation}
T_c(X_\tau(x))=T_c(x),\qquad J_xT_c(x)b_c(x)=0,\qquad T_c(y)=y.\label{eq:terminal}
\end{equation}
Transport time $\tau$ is distinct from video time. The auxiliary transport evolves the transport state while keeping the observation history and actions fixed. We learn its endpoint directly. Equation~\ref{eq:terminal} is inherited from BTM. We develop an action-conditioned terminal-map predictor and address two challenges in one step video prediction: directly supervising the map from pure source noise and reducing error accumulation under self-generated histories. 

\begin{figure}[!htb]
\centering\includegraphics[width=.76\linewidth]{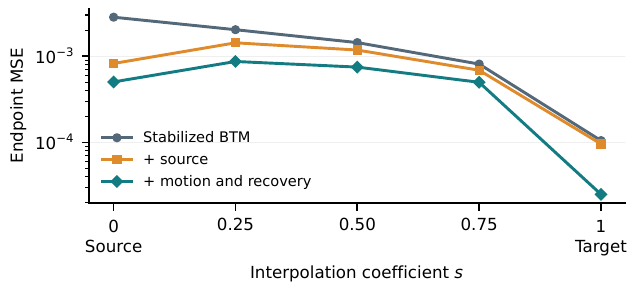}
\caption{\textbf{Source anchoring improves prediction from initial noise.} Measured Can endpoint MSE on 90 fixed windows and four shared source draws per window. The stabilized map has low error at the target boundary ($s=1$) but much higher error at the source ($s=0$). Source anchoring reduces source MSE by $71\%$. The continuation with motion supervision and generated-history recovery reduces it further. Points denote measured input locations. Connecting lines guide the eye. These checkpoints precede spatial training.}
\label{fig:terminal}
\end{figure}

\subsection{Pair-Based Terminal-Map Training}
Sample $s\sim\mathcal U[0,1]$ and form a pair of noise and a ground-truth future chunk,
\begin{equation}
x_s=(1-s)z+sy,\qquad d=y-z,\qquad p=T_\theta(x_s,c).
\end{equation}
Following BTM, we construct a target using a Jacobian-vector product (JVP) and apply stop-gradient,
\begin{equation}
u=\sg\!\left[p+\eta J_xT_\theta(x_s,c)d\right],\qquad \eta=0.25.\label{eq:jvp}
\end{equation}
Here $\sg$ stops reverse-mode gradients. The JVP differentiates only the future input $x_s$. $h$ and $a$ are fixed. The displacement $d = y - z$ defines the JVP direction along the noise–data interpolant, without requiring a separate velocity network. Although $u$ contains a derivative, the stopped target avoids backpropagating through that derivative during the parameter update.

The transport residual uses detached per-example weights, boundary and near-boundary reconstruction, and an output-gradient balance, yielding $\Lbtm$ (Equations~\ref{eq:transport} through~\ref{eq:btm} in Appendix~\ref{app:implementation}). These weighting and boundary terms stabilize training.

\subsection{Source and Endpoint Supervision}
The one-step prediction from initial noise is $p_0=T_\theta(z,c)$. Our fixed-window diagnostic shows that accurate reconstruction at the data boundary does not ensure an accurate $p_0$ (Figure~\ref{fig:terminal}). We therefore add source anchoring,
\begin{equation}
\Lsrc=\E\normD{p_0-y}.\label{eq:source}
\end{equation}
Source anchoring directly supervises this prediction. It penalizes conditional-mean error and variation induced by source noise. This is useful for paired manipulation prediction but may suppress valid conditional diversity. Appendix~\ref{app:diagnostics} gives the exact conditional-variance decomposition in Equation~\ref{eq:variance}.

\begin{figure}[t]
\centering
\includegraphics[width=\linewidth]{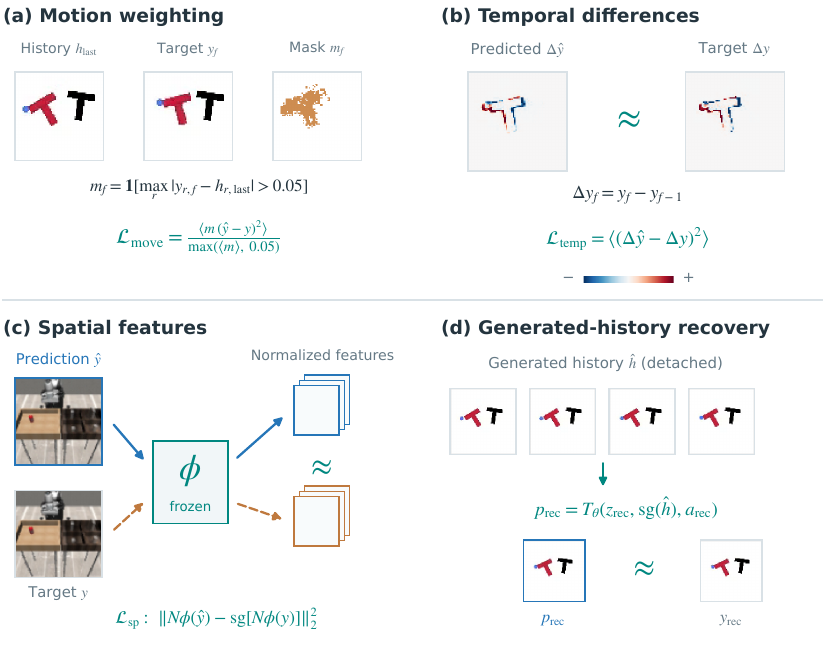}
\caption{\textbf{Endpoint supervision.} (a) A target-defined mask weights changed pixels, normalized per example. The notation $\langle\cdot\rangle$ denotes averaging over tensor coordinates. (b) Temporal differences are compared within a future chunk. Color shows channel-averaged signed RGB differences on a shared scale. (c) A frozen VGG16 encoder matches three levels of normalized features, detaching only target features. (d) A detached generated prefix conditions a new prediction using actions and targets from the dataset. Fixed PushT crops and saved Can rollout frames illustrate the losses, rather than a captured training batch. Appendix~\ref{app:implementation} gives exact definitions and schedules.}
\label{fig:training-signals}
\end{figure}

\paragraph{Endpoint regularization.}
Uniform pixel error can be dominated by static backgrounds. We emphasize pixels that change relative to the last observation, match within-chunk temporal differences, and softly penalize out-of-range RGB values,
\begin{equation}
\mathcal A(p,y)=0.1\mathcal L_{\mathrm{move}}+0.05\mathcal L_{\mathrm{temp}}+0.01\mathcal L_{\mathrm{range}}.\label{eq:aux}
\end{equation}
Figure~\ref{fig:training-signals} illustrates these endpoint and recovery signals. The moving mask is target-defined. The final training stage matches three levels of frozen ImageNet VGG16 features at $p_0$ \citep{simonyan2015vgg}, with detached target features and channel normalization (Equation~\ref{eq:spatial}, Appendix~\ref{app:implementation}).
We reuse the source prediction $p_0$ to compute this loss and discard the feature encoder at inference.

\paragraph{Generated-history recovery.}
Every fourth update, a detached generated prefix replaces the visual history for an additional prediction using actions and targets from the dataset. The weight is $0.25$ on active updates, with a depth curriculum. We condition the model on its own detached predictions while retaining the corresponding ground-truth future as the target, training it to tolerate accumulated prediction errors (Equation~\ref{eq:recovery}, Appendix~\ref{app:implementation}).

After the source-loss weight reaches its final value, the pixel-space training objective is
\begin{equation}
\mathcal L_{\mathrm{RGB}}=\Lbtm+\Lsrc+r_k\mathcal A(p_0,y)+\mathcal R+\alpha_{k_s}\mathcal L_{\mathrm{sp}},\label{eq:full}
\end{equation}
where $r_k=\min(k/1000,1)$ and $\alpha_{k_s}=0.05\min(k_s/1000,1)$. Training introduces source anchoring, motion supervision and generated-history recovery, and spatial supervision in successive stages. The spatial stage retains the mature recovery schedule. Algorithm~\ref{alg:train} summarizes the computation.

\FloatBarrier
\subsection{Latent-Space Video Prediction}\label{sec:native-method}
For $256\times256$ RT-1 and Bridge-V2 video, one forward pass predicts one SD3 VAE latent frame \citep{esser2024sd3}. Transport and source losses operate in the latent space and while VGG perceptual supervision operates on decoded frames.  We retain conditional drifting in frozen DINOv3 and latent features \citep{simeoni2025dinov3,lu2026driftworld}, using $K=8$ samples and coefficient $0.01$, and add the same term at generated-history recovery endpoints. Equation~\ref{eq:native} and Appendix~\ref{app:native} give the exact hybrid objective. Its generated samples are used only in training. Inference requires one forward pass per frame.

%% file: sections/experiments.tex
\section{Experiments}\label{sec:experiments}
We evaluate BWM through the cost of learning its predictor, the accuracy of its autoregressive predictions, and its ability to estimate policy performance at low inference cost. A separate latent-video study tests whether feature supervision under generated histories improves perceptual prediction. 

\subsection{Experimental Setup}\label{sec:setup}
\paragraph{RGB dynamics.}
We evaluate PushT \citep{chi2023diffusionpolicy} and Robomimic Can \citep{mandlekar2021robomimic} at $96\times96$ resolution. Each 64-frame evaluation starts from four frames and predicts the remaining 60 frames autoregressively from actions. This benchmark includes 388 PushT and 30 Can episodes. Full-episode evaluation includes 400 and 30 episodes respectively. Pixel MSE measures paired prediction error, while LPIPS, SSIM, and PSNR characterize appearance. Metrics are averaged within each episode and then across episodes. Appendix~\ref{app:evaluation} gives the metric conventions and episode-bootstrap intervals.

\paragraph{Baselines and training budgets.}
Local DriftWorld and MSE regression use the corresponding action-conditioned backbone. The MSE U-Net is trained directly with squared prediction error. The AVDC-derived control uses a time-conditioned U-Net with an adapted diffusion recipe. Local baselines use five capped epochs on our split. Released GPC and DriftWorld checkpoints are additional references. All reported scores are obtained under our evaluation protocol.

\paragraph{Policy evaluation and timing.}
Following DriftWorld's PushT protocol, we evaluate seven frozen diffusion-policy checkpoints \citep{chi2023diffusionpolicy,lu2026driftworld}. Each policy is evaluated on 300 trials paired with simulator trials. We measure the mean absolute difference between the seven policy-level mean IoUs and the Pearson correlation across those means. These metrics assess how closely predicted policy performance agrees with simulator performance. Model timings measure batch-one inference after warmup on an A800 and exclude the policy and the pose-estimation networks used to score predictions. All RGB predictors produce four frames per prediction in this comparison. Table~\ref{tab:policy} reports their numbers of network evaluations.

\subsection{Training Efficiency}\label{sec:efficiency}
\begin{figure}[!htb]
\centering\includegraphics[width=\linewidth]{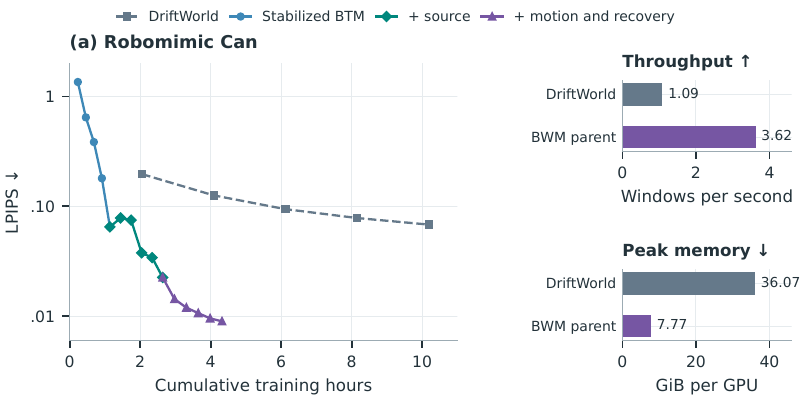}
\caption{\textbf{Can prediction quality and training resources.} Left: evaluated checkpoints against cumulative training-loop time, counting all preceding BWM stages. The BWM curve ends at the checkpoint trained with motion supervision and generated-history recovery, before spatial supervision. Right: complete-update profiling on two A800 GPUs, with one conditioning window per GPU. Throughput counts these primary windows, including the cost of auxiliary recovery predictions. Peak memory is allocated memory per GPU. Both models use one call at inference.}
\label{fig:efficiency}
\end{figure}

We compare BWM with local DriftWorld, which already has the same single-pass prediction interface, through profiling of complete training steps and prediction quality over cumulative training time.
Figure~\ref{fig:efficiency} examines complete-update resources and prediction quality over cumulative training time. 

\paragraph{Training-step cost.}
The BWM training procedure has a lower measured update cost even after accounting for recovery predictions. At equal batches of windows, BWM achieves $3.3\times$ higher throughput with $78\%$ lower peak allocated memory (Figure~\ref{fig:efficiency}). Appendix~\ref{app:resources} gives the absolute throughput and memory measurements.

% \paragraph{Quality within a training budget.}
% The lower update cost is accompanied by better Can prediction within the time budget. At the last evaluated checkpoints within five cumulative hours, BWM has LPIPS $0.0090$ and local DriftWorld $0.1257$; DriftWorld reaches $0.0682$ at $10.19$ hours. BWM's $4.32$ hours include its stabilized, source-anchored, and motion/recovery stages, with no clock reset between stages. This establishes earlier attainment of accurate predictions in these runs, without inferring an unobserved threshold crossing or a universal convergence rate. The final Can spatial stage adds $12.07$ training hours and supplies a different checkpoint for the prediction tables below. PushT also inherits a longer history; the Can curve does not establish a lower total training budget for those final models (Appendix~\ref{app:evaluation}).

\subsection{Policy Evaluation}\label{sec:policy}
\begin{figure}[t]
\centering
\includegraphics[width=\linewidth]{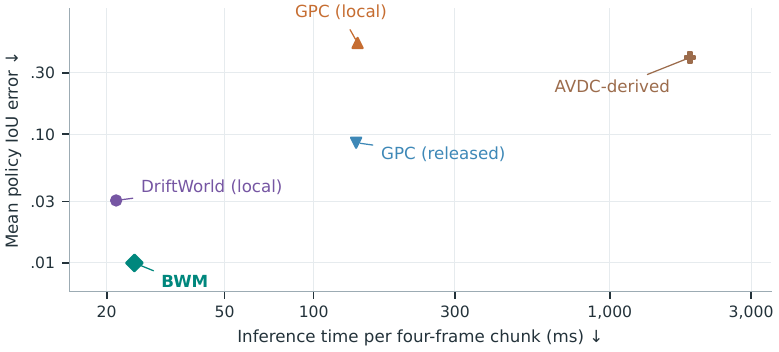}
\caption{\textbf{Policy-score accuracy versus inference speed on PushT.} Error is the mean absolute difference between model and simulator policy-level mean IoUs. Latency is for four predicted frames. Training histories and sampler calls are given in Table~\ref{tab:policy}. Timings exclude policy execution and reward estimation.}
\label{fig:frontier}
\end{figure}

Accurate image prediction is useful for policy evaluation when the imagined outcomes preserve differences in policy performance. We test this directly by comparing predicted policy scores with simulator scores. Figure~\ref{fig:frontier} compares policy-score estimation error against per-chunk inference latency.

\input{tables/policy}

BWM achieves a mean absolute policy evaluation error of $0.0100$, compared with $0.0306$ for local DriftWorld and $0.0203$ for its released checkpoint. BWM and released DriftWorld have similar correlations across policy means, but BWM more closely reproduces their absolute levels (Table~\ref{tab:policy}). The result shows why absolute error provides information beyond correlation when evaluating a predictive simulator.

This accuracy is obtained at $24.8$\,ms per four-frame prediction, close to the other one-step predictors. Against released GPC, BWM has lower policy-score error while requiring $5.6\times$ less inference time. The comparison therefore supports accurate policy-performance estimation without iterative sampling at inference. Table~\ref{tab:policy} reports the evaluated generative models. The AVDC-derived control underestimates policy IoU, as detailed in Appendix~\ref{app:policy}.

\paragraph{Model-based planning.}
Planning provides a separate test of whether predictions improve executed actions. With fixed GPC-RANK proposals \citep{qi2026gpc}, BWM achieves mean executed IoU $0.721$, compared with $0.709$ for local DriftWorld and $0.695$ for the MSE U-Net. The paired difference between BWM and DriftWorld is $+0.012$, with a $95\%$ bootstrap interval of $[-0.047,0.071]$. BWM achieves the highest mean executed IoU, while its clearest downstream advantage is more accurate policy-score estimation. 

\subsection{Autoregressive Prediction}\label{sec:rgb}
\input{tables/rgb}
We next examine whether accurate policy scores are accompanied by faithful visual dynamics over repeated predictions. On the 64-frame benchmark, the final BWM model reduces pixel MSE by $68\%$ on PushT and $57\%$ on Can relative to same-backbone regression. Moving-region error also decreases, and perceptual quality improves on both datasets (Table~\ref{tab:rgb}). These complementary measurements show that the improvement extends to changing robot and object pixels. Full-episode evaluation retains lower MSE than the local baselines (Appendix~\ref{app:evaluation}).

\begin{figure}[t]
\centering\includegraphics[width=\linewidth]{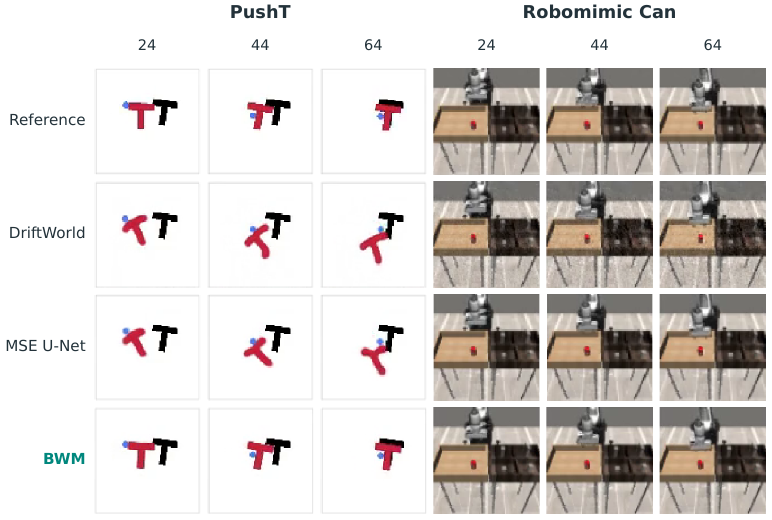}
\caption{\textbf{Autoregressive RGB predictions.} Reference and predicted frames at positions 24, 44, and 64 in two selected evaluation sequences, with the same actions and times across rows. BWM uses the final spatial continuation. Local baselines use their five-epoch fits. Appendix~\ref{app:qualitative} gives sample selection and additional baseline rows.}
\label{fig:rgb}
\end{figure}

The qualitative rollouts illustrate the structure behind these aggregate errors. In Figure~\ref{fig:rgb}, BWM more closely preserves the moving object's pose in PushT and the configuration of the robot arm and object in Can. Appendix~\ref{app:qualitative} provides additional baseline rows and describes sample selection.

The released DriftWorld checkpoint has lower PushT pixel MSE and LPIPS than BWM under our evaluation (Appendix~\ref{app:released}). It has a much longer training history and was not trained with our episode split. Thus the final RGB comparisons establish improvements over our local controls, while the released-reference comparison identifies lower policy-score error as BWM's advantage. Visual fidelity and policy-score agreement assess different aspects of world-model quality.

% \paragraph{Matched-exposure control.}\label{sec:fresh}
% A fresh BWM fit uses the same five capped epochs and clean-window exposure as the local baselines, with a predeclared final EMA checkpoint. On PushT it reduces pixel MSE by $30\%$ relative to regression, but worsens LPIPS by $22\%$; paired episode-bootstrap intervals separate both differences from zero. On Can, the small 60-frame MSE advantage has a paired interval spanning zero and reverses over full episodes, where MSE is $14\%$ worse. Fixed examples show diffuse object shapes and loss of arm/object structure (Appendix~\ref{app:fresh}). The final staged model's joint pixel and perceptual gains are therefore not reproduced by fresh training at this exposure. Additional recovery supervision and different optimizers also mean that matched clean exposure does not isolate the objective at equal computation.

\subsection{Ablation Analysis}\label{sec:ablation}
The source diagnostic tests whether boundary reconstruction is an adequate proxy for prediction from Gaussian noise. On fixed Can windows, the stabilized map has small boundary error but substantially larger source error (Figure~\ref{fig:terminal}). Source anchoring reduces source MSE from $0.002831$ to $0.000823$, a $71\%$ reduction. This directly supports supervising the prediction from Gaussian noise used at inference.

The subsequent training stages improve the quality of repeated predictions. Adding motion supervision and generated-history recovery reduces full-episode Can MSE from $0.0940$ to $0.0112$, with a smaller improvement on PushT. Continued spatial supervision further improves LPIPS on both datasets. 

\subsection{Real-World Video Prediction}\label{sec:native}
We examine whether extending feature supervision to generated histories improves a fixed BWM checkpoint on RT-1 \citep{brohan2022rt1} and Bridge-V2 \citep{walke2023bridge}. The models use $256\times256$ videos and latents encoded by the frozen SD3 VAE. Each evaluation window starts from four context frames and predicts eight frames autoregressively. Context is restored at the start of the next window. References are VAE reconstructions. This evaluates conditioned video prediction separately from the RGB benchmark's free-running rollouts.

\begin{figure}[t]
\centering\includegraphics[width=\linewidth]{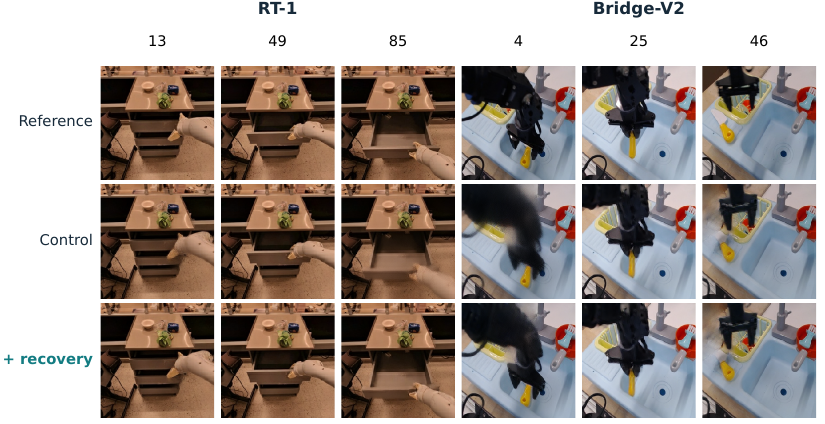}
\caption{\textbf{Native-video prediction with generated-history feature supervision.} Reference, control, and recovery-refined predictions at matched positions in selected RT-1 and Bridge-V2 episodes. References are VAE reconstructions. Each eight-frame prediction window starts from four frames. Separated columns span several such windows. Residual gripper blur remains visible.}
\label{fig:native}
\end{figure}
\input{tables/native}

Both fine-tuning conditions begin from the same epoch-70 EMA checkpoint trained on the full dataset, retain the history feature loss, and use fresh optimizers. The added term applies feature drifting to predictions conditioned on generated histories. The additional training comprises 1,000 updates on RT-1 and 500 on Bridge-V2. Because the control already has history feature supervision, this comparison tests the benefit of extending that supervision to generated contexts.

The resulting improvement is clearest in perceptual prediction. LPIPS decreases by $34\%$ on RT-1 and $24\%$ on Bridge-V2 relative to their controls. FID and FVD also improve on both datasets, while PSNR decreases slightly (Table~\ref{tab:native}). Thus the added supervision improves feature-based measures of frame appearance and video distribution without a corresponding pixelwise improvement. Figure~\ref{fig:native} provides matched qualitative examples, including the residual gripper blur.

% This study evaluates additional training from the same initial checkpoint, with checkpoints selected through validation review. (Appendix~\ref{app:native-subset}).

% A separate paired-feature recovery variant is evaluated on 213 Bridge-V2 episodes (Appendix~\ref{app:paired-native}). It improves LPIPS by $4.5\%$. That variant retains history drifting and replaces the added drifting loss under generated histories. 

%% file: tables/policy.tex
\begin{table}[t]
\centering\small\setlength{\tabcolsep}{5pt}
\caption{\textbf{PushT policy evaluation and inference cost.} Error and correlation use seven policy means, each from 300 paired trials. Calls and latency refer to four predicted frames. Local baselines are five-epoch fits. }
\label{tab:policy}
\begin{tabular}{@{}llrrrr@{}}\toprule
World model & Training & Calls & ms $\downarrow$ & IoU error $\downarrow$ & $r$ $\uparrow$\\\midrule
% MSE U-Net & Local & 1 & 21.8 & .0121 & .9150\\
DriftWorld & Local & 1 & \textbf{21.5} & .0306 & .8565\\
GPC & Local & 12 & 140.7 & .5127 & .3659\\
AVDC & Local & 100 & 1866.2 & .3951 & -.0801\\
\midrule
DriftWorld & Released & 1 & \textbf{21.5} & .0203 & .9255\\
GPC & Released & 12 & 139.1 & .0860 & .8438\\
\midrule
\textbf{BWM} & Local & 1 & 24.8 & \textbf{.0100} & \textbf{.9280}\\
\bottomrule\end{tabular}
\end{table}

%% file: tables/rgb.tex
\begin{table}[t]
\centering\small\setlength{\tabcolsep}{4.8pt}
\caption{\textbf{RGB prediction on episode-held-out splits.} MSE, moving-region MSE, and LPIPS for 60 autoregressively predicted frames after the initial context. All metrics are lower-is-better. BWM is the final spatial-stage model. Baselines are local five-epoch fits. These rows compare different total training budgets.}
\label{tab:rgb}
\begin{tabular}{@{}lrrrrrr@{}}\toprule
&\multicolumn{3}{c}{PushT}&\multicolumn{3}{c}{Robomimic Can}\\
\cmidrule(lr){2-4}\cmidrule(lr){5-7}
Model&MSE&Moving MSE&LPIPS&MSE&Moving MSE&LPIPS\\\midrule
DriftWorld&.0314&.4093&.0790&.0104&.0769&.0682\\
MSE U-Net&.0303&.4008&.0804&.0051&.0544&.0104\\
GPC&.0630&1.4241&.1609&.0276&.3103&.0589\\
AVDC&.0684&1.1840&.1860&.0400&.2644&.1194\\
\midrule
\textbf{BWM}&\textbf{.0098}&\textbf{.1247}&\textbf{.0280}&\textbf{.0022}&\textbf{.0242}&\textbf{.0056}\\
\bottomrule\end{tabular}
\end{table}

%% file: tables/native.tex
\begin{table}[t]
\centering\small\setlength{\tabcolsep}{5.5pt}
\caption{\textbf{Feature supervision under generated histories.} Full-data BWM parents and matched-length continuations. RT-1 adds 1,000 updates. Bridge-V2 adds 500. The control retains clean-history feature drifting. Recovery adds it at generated-history endpoints. Both continuations use the same parent and one training seed.}
\label{tab:native}
\begin{tabular}{@{}llrrrr@{}}\toprule
Dataset&Model&PSNR $\uparrow$&LPIPS $\downarrow$&FID $\downarrow$&FVD $\downarrow$\\\midrule
RT-1&Parent&\textbf{27.35}&.0917&19.77&167.50\\
&Control&27.34&.0933&20.56&172.89\\
&$+$ recovery&27.27&\textbf{.0612}&\textbf{8.05}&\textbf{104.91}\\\midrule
Bridge-V2&Parent&25.00&.1113&31.86&175.53\\
&Control&\textbf{25.01}&.1106&30.98&171.56\\
&$+$ recovery&24.95&\textbf{.0841}&\textbf{21.34}&\textbf{117.20}\\
\bottomrule\end{tabular}
\end{table}

%% file: sections/discussion.tex
\section{Discussion and Conclusion}
Beckmann World Models learn action-conditioned futures as terminal maps. The central empirical finding is that a pair-based RGB training procedure can retain single-pass inference while reaching accurate Can predictions earlier, with higher training throughput and lower peak memory than local DriftWorld. Source anchoring and endpoint supervision also produce accurate policy-score predictions at a similar inference cost to the other one-step models. On real robotics videos, feature supervision under generated histories improves perceptual quality without adding inference steps. 
% Together, these results connect the choice of generative training procedure to both the cost of learning a visual simulator and the utility of its predictions.

%% file: sections/statements.tex
\section*{Reproducibility Statement}
Appendices~\ref{app:implementation} through~\ref{app:fresh} specify the losses, tensor alignment, training lineage, evaluation protocols, and additional results. Episode-bootstrap intervals quantify evaluation-cohort uncertainty. The reported experiments use one training seed. 
\section*{Ethics Statement}
This work studies visual prediction for manipulation using existing datasets. Model-generated outcomes can be inaccurate, and the reported evaluations do not establish suitability for safety-critical robotic control.
\section*{Statement of AI Use}
Generative AI tools assisted with experiment execution, and debugging. Quantitative results come from model evaluation records, and empirical visualizations use dataset observations and model outputs. The authors are responsible for the final paper.

%% file: appendices/implementation.tex
\section{Implementation Details}
\label{app:implementation}

\subsection{Gradient Balancing and Optimization}
Let $w$ denote parameters of the output spatial projection. Define
\begin{equation}
 g_{\mathrm{tr}}=\nabla_w\mathcal L_{\mathrm{tr}},\qquad
 g_{\mathrm{bd}}=\nabla_w(\mathcal L_{\mathrm{bd}}+\mathcal L_{\mathrm{near}}).
\end{equation}
We average each gradient vector across distributed ranks before taking its Euclidean norm. When both norms exceed $10^{-8}$, the detached balance is
\begin{equation}
 \lambda=\sg\!\left(\operatorname{clip}\left(
 \frac{\lVert g_{\mathrm{bd}}\rVert_2}{\lVert g_{\mathrm{tr}}\rVert_2},
 10^{-3},10^3\right)\right).
\end{equation}
When either finite norm is at most $10^{-8}$, the coefficient is one. Nonfinite gradients cause a failure rather than a silent substitution. The fallback prevents an initially zero boundary gradient from disabling transport. Source and endpoint auxiliary terms are not inserted into this original balancing group.

The stabilized map uses Muon for hidden weights and AdamW for output and auxiliary parameters. The original target learning rates are $10^{-3}$ and $10^{-4}$. The completed motion-aware continuation uses one quarter of these rates. Warmup lasts 500 updates from one percent of the target rate. The global gradient norm is clipped at two, and EMA decay is $0.999$. Training and the full-update comparison use full precision. Each continuation explicitly initializes from its parent's EMA and resets the continuation's optimizer and scheduler. Resuming the same continuation instead restores its own model, optimizer, scheduler, EMA, RNG states, sampler position, and update clock.

The matched DriftWorld backbone does not imply identical optimization. Its training runs use AdamW, smaller window batches, and 16 generated negatives per PushT condition or 32 per Robomimic Can condition. Kernel temperatures are $0.02$, $0.05$, and $0.2$. These differences are part of the complete implementation comparison and prevent attributing every speed or quality difference to a single operation.

\subsection{Exact Endpoint Regularizers}
For example $i$, let $y_{i,0}$ be the last observed frame before the target chunk. The target-defined moving mask is
\begin{equation}
 m_{i,f,u,v}=\mathbf 1\!\left[
 \max_{r\in\{1,2,3\}}|y_{i,r,f,u,v}-y_{i,r,0,u,v}|>0.05\right].
\end{equation}
The RGB channel index is $r$, future position is $f$, and spatial coordinates are $(u,v)$. Broadcast the mask over channels and let $\operatorname{mean}$ average the indicated tensor coordinates. The training loss is
\begin{equation}
 \mathcal L_{\mathrm{move}}=\frac1B\sum_i
 \frac{\operatorname{mean}\!\left[m_i(p_i-y_i)^2\right]}
 {\max(\operatorname{mean}(m_i),0.05)}.
\end{equation}
The floor limits the additional per-coordinate weight induced by the coefficient $0.1$ to at most two. The other terms are
\begin{align}
 \mathcal L_{\mathrm{temp}}&=\operatorname{mean}_{i,r,f=2:F,u,v}
 \left[\big((p_{i,r,f,u,v}-p_{i,r,f-1,u,v})
 -(y_{i,r,f,u,v}-y_{i,r,f-1,u,v})\big)^2\right],\\
 \mathcal L_{\mathrm{range}}&=\operatorname{mean}\left[\max(|p|-1,0)^2\right].
\end{align}
The temporal term concerns within-chunk differences. It does not force a repaired endpoint to preserve the offset of an erroneous generated preceding frame.

The recovery branch uses a separate valid 16-frame training sequence. Robomimic Can generates between one and seven two-frame prefix chunks before a two-frame endpoint. PushT generates between one and three four-frame prefix chunks and alternates four-frame and three-frame endpoints. The depth curriculum expands over 1,000 completed updates. Prefix generation is detached, raw, and conditioned on aligned dataset actions. No blend with the true history or inference clamp is introduced. The branch is active every fourth update, with weight $0.25$ on active updates. It is not multiplied by four to compensate for its frequency. All losses use one final backward pass and one optimizer update.

The clean-window dataset retains short episodes that cannot support a 16-frame sequence. The recovery sequence is obtained from a separate training-only valid-sequence index. No sample crosses an episode boundary. This regularizer does not create labeled counterfactual simulator transitions.

\subsection{Frozen Spatial Endpoint Features}
Spatial supervision uses ImageNet-pretrained VGG16 convolutional weights \citep{simonyan2015vgg} and no classifier. Each predicted video tensor is reshaped from $[B,3,F,96,96]$ to $[BF,3,96,96]$, mapped from $[-1,1]$ to $[0,1]$, and normalized by the standard ImageNet channel statistics. We extract relu1\_2, relu2\_2, and relu3\_3 at their native spatial resolutions. For feature tensor $v\in\mathbb R^{C\times U\times V}$, channel normalization is
\begin{equation}
 \bar v_{:,u,v}=\frac{v_{:,u,v}}{\lVert v_{:,u,v}\rVert_2+10^{-6}}.
\end{equation}
The loss at each level is the mean squared Euclidean distance between normalized channel vectors, summed over channels and averaged over batch, frame, and spatial locations. The three level losses receive equal weight. Target features are computed without gradients. Predictions are not clipped before feature extraction, so out-of-range outputs still contribute gradients.

The spatial coefficient ramps linearly from zero to $0.05$ over 1,000 child updates. The VGG module is frozen and is stored outside the world model, optimizer, EMA, and checkpoint state. The source prediction already required by $\Lsrc$ is reused, so spatial supervision introduces no additional world-model forward pass. Training fails if the declared feature asset or checksum is unavailable rather than silently substituting random features. At inference the feature module is absent.

\paragraph{Continuation schedule.} The base map first learns $\Lbtm$. Source supervision ramps to one over 500 updates. Motion supervision and generated-history recovery terms ramp over 1,000 updates from the source-anchored EMA. The final spatial stage continues the motion-aware EMA, ramps only its feature coefficient, and retains the parent recovery clock.

\subsection{Stabilized Transport Objective}
We stabilize the transport residual using a detached per-example weight and boundary supervision:
\begin{align}
e_i&=\normD{p_i-u_i},\qquad
\mathcal L_{\mathrm{tr}}=\frac1B\sum_i\frac{e_i}{\sg(e_i)+0.01},\label{eq:transport}\\
\mathcal L_{\mathrm{bd}}&=\E\normD{T_\theta(y,c)-y},\qquad
\mathcal L_{\mathrm{near}}=\E\!\left[\mathbf1_{s>0.8}\normD{p-y}\right],\\
\Lbtm&=\lambda\mathcal L_{\mathrm{tr}}+\mathcal L_{\mathrm{bd}}+\mathcal L_{\mathrm{near}}.\label{eq:btm}
\end{align}
The detached $\lambda$ balances the transport and boundary gradient norms on the output projection. These stabilizers and the optimizer recipe are empirical adaptations, not consequences of the transport theorem. Appendix~\ref{app:implementation} gives their exact implementation.

\subsection{Spatial Feature Supervision}
The moving mask comes from the target, so prediction error cannot define where supervision is applied. A final continuation adds frozen ImageNet VGG16 features \citep{simonyan2015vgg}. With $N$ denoting channel normalization at each spatial location,
\begin{equation}
\mathcal L_{\mathrm{sp}}=\frac13\sum_{\ell=1}^{3}
\operatorname{mean}_{b,f,u,v}\left\|
N(\phi_\ell(p_0)_{bfuv})-N(\phi_\ell(y)_{bfuv})\right\|_2^2.\label{eq:spatial}
\end{equation}
The three levels are relu1\_2, relu2\_2, and relu3\_3. The target features are detached. Gradients pass through the frozen encoder to the prediction. Reusing $p_0$ adds no world-model forward pass, and the encoder is discarded at inference.

\subsection{Training Algorithm}
\begin{algorithm}[H]
\caption{RGB terminal-map training and inference}\label{alg:train}
\begin{algorithmic}[1]
\Require Training windows $(h,a,y)$, active training stage, and EMA parameters $\bar\theta$
\State Sample $z\sim\mathcal N(0,1.35^2I)$ and $s\sim\mathcal U[0,1]$
\State $x_s\gets(1-s)z+sy$, $d\gets y-z$, and $c\gets(h,a)$
\State Compute $(p,J_xT_\theta(x_s,c)d)$ with a JVP holding $c$ fixed
\State Construct $u$ and $\Lbtm$ by Equations~\ref{eq:jvp} and~\ref{eq:transport} through~\ref{eq:btm}
\If{source supervision is active}
\State $p_0\gets T_\theta(z,c)$ and add $\Lsrc$ and the active endpoint losses
\EndIf
\If{recovery is active on this update}
\State Generate a detached prefix on a valid training sequence
\State Predict its next endpoint and add $\mathcal R$ from Equation~\ref{eq:recovery}
\EndIf
\State Backpropagate the combined loss once and update $\theta$ and its EMA $\bar\theta$
\Statex \textbf{Inference:} sample $z$ and return $T_{\bar\theta}(z,h,a)$
\end{algorithmic}
\end{algorithm}

\subsection{Generated-History Endpoint Objective}
\paragraph{Recovery from generated histories.}
Every fourth update, a detached model-generated prefix replaces the visual history for a separate training endpoint. The aligned actions and target are taken from the dataset. If $p_{\mathrm{rec}}=T_\theta(z_{\mathrm{rec}},\sg(\widehat h),a_{\mathrm{rec}})$, then
\begin{equation}
\mathcal R=0.25r_k I_k\left[\normD{p_{\mathrm{rec}}-y_{\mathrm{rec}}}+\mathcal A(p_{\mathrm{rec}},y_{\mathrm{rec}})\right],\quad
I_k=\mathbf1_{(k+1)\bmod4=0}.\label{eq:recovery}
\end{equation}
The prefix exposes the predictor to its own context errors. Stopping it prevents backpropagation through a long rollout. The dataset endpoint remains the recovery target, not the physical consequence of a counterfactual generated state.

%% file: appendices/evaluation.tex
\clearpage
\section{Evaluation, Training Lineage, and Additional Results}\label{app:evaluation}
\subsection{RGB Evaluation}
We use EMA weights and evaluation seed 314. For the 64-frame protocol, an episode must contain the four-frame context and 60 future frames. This leaves 388 PushT episodes and 30 Can episodes. Full-episode evaluation uses all 400 PushT episodes and the same 30 Can episodes. Predictions are fed back without an inference clamp. Targets and controls are taken from the dataset episode, and a sequence never crosses an episode boundary.

Pixel MSE is computed on raw $[-1,1]$ RGB predictions. LPIPS \citep{zhang2018lpips} uses clipped images. PSNR and SSIM \citep{wang2004ssim} use the clipped, quantized image convention of the evaluator. Moving-region MSE selects target pixels that differ from the last context frame by more than $0.05$ in at least one channel. The metric is averaged within episodes and then equally across episodes. This evaluation statistic is distinct from the floored motion-loss normalization used during training.

Table~\ref{tab:rgb-complete} reports the complete final BWM metrics. We recomputed $95\%$ percentile intervals by resampling complete episodes 10,000 times (resampling seed 943), separately for each task and protocol. These intervals describe evaluation-cohort variation for one trained model, not uncertainty across training seeds. 
\input{tables/rgb_complete}

% \subsection{Training History}\label{app:lineage}
% Can trains from scratch through 5,000 stabilized-map updates, 5,000 source updates, and 5,000 motion/recovery updates before the spatial stage. PushT inherits a longer initial training history; its motion-aware parent contains 182,978 cumulative updates. Spatial training continues the same child state from epochs 5 through 30, rather than restarting at epoch 5. Parent-to-child recipe changes initialize from the parent's EMA and start the new stage's optimizer; ordinary continuation restores the existing optimizer and schedule.

% \begin{table}[ht]\centering\small\setlength{\tabcolsep}{6pt}
% \caption{\textbf{Final RGB training lineage.} BWM's inherited updates are counted in the total. Local DriftWorld and MSE U-Net are five-epoch fits from scratch. Update counts alone do not equate exposure when batch sizes differ.}\label{tab:lineage}
% \begin{tabular}{@{}llrrr@{}}\toprule
% Task&Model/stage&Inherited&Added&Total\\\midrule
% PushT&Spatial, 5 epochs&182,978&19,675&202,653\\
% &Spatial, 30 epochs&182,978&118,050&301,028\\
% &Local DriftWorld / MSE U-Net&0&39,350&39,350\\\midrule
% Can&Spatial, 5 epochs&15,000&5,000&20,000\\
% &Spatial, 30 epochs&15,000&30,000&45,000\\
% &Local DriftWorld / MSE U-Net&0&20,000&20,000\\
% \bottomrule\end{tabular}\end{table}

The cumulative Can curve sums logged training-loop intervals across the stabilized-map, source-anchoring, and motion and recovery runs. The respective increments are 1.14, 1.50, and 1.68 hours, totaling 4.32 hours on two A800 GPUs. Local DriftWorld uses 10.19 hours on the same GPU count. The curves show evaluated checkpoints without smoothing or interpolation of an unobserved threshold crossing. They exclude evaluation waits and data preparation.

\subsection{Training-Cost Comparison}
DriftWorld is the key comparison for training efficiency because it already shares BWM's one-call prediction interface. We evaluate two distinct consequences of replacing its conditional generated population with pair-based RGB terminal-map training: the resources needed for complete updates, and the quality reached over cumulative wall-clock time. Figure~\ref{fig:efficiency} presents both. Neither can be inferred from the number of inference calls alone.

\subsection{Resource and Inference Measurements}\label{app:resources}
The complete-update comparison uses two A800 80GB GPUs and one primary conditioning window per GPU. It includes the forward and JVP computations, all active losses, backward pass, distributed synchronization, clipping, optimizer, scheduler, and EMA update. The mature recovery schedule is exercised across its prefix depths. Peak memory is the maximum PyTorch allocated memory across ranks, including model, optimizer, and EMA state. The spatial encoder is absent from this particular measurement. Throughput counts primary windows rather than treating auxiliary generated samples as additional independent training examples.

\begin{table}[ht]\centering\small\setlength{\tabcolsep}{6pt}
\caption{\textbf{Complete-update cost for the Can RGB recipes.} Equal batches of training windows in full precision.}\label{tab:resources}
\begin{tabular}{@{}lrrrr@{}}\toprule
Recipe&Mean s&95th percentile s&Windows $\mathrm{s}^{-1}$&Peak GiB\\\midrule
Local DriftWorld&1.841&1.843&1.09&36.07\\
Source-anchored map&.496&.509&4.03&6.12\\
Motion and recovery&.553&.739&3.62&7.77\\
\bottomrule\end{tabular}\end{table}

Inference timings measure batch-one model execution on an A800 after warmup, with the model resident in GPU memory. Local baseline, final BWM, and newly evaluated released GPC and AVDC-derived profiles were measured in separate runs, so small latency differences among one-call models should not be interpreted as an architectural speed advantage. All reported latencies are means: BWM averages 40 synchronized measurements across eight contexts (24.82\,ms, median 24.69\,ms), local baselines use mean inference time over 20 iterations, and released GPC and AVDC each average 20 measurements. BWM measures the terminal-map call with preallocated noise. Baseline profiles include sampler-side noise creation. The separately measured profiles therefore compare model execution cost, rather than a common end-to-end serving stack. GPU exclusivity was not independently verified for the final BWM profile. Native VAE decoding and the policy and reward networks are outside the RGB generator comparison.

\subsection{Policy Evaluation and Planning}\label{app:policy}
For policy evaluation, the seven frozen diffusion-policy checkpoints are epochs 50, 100, 150, 200, 250, 300, and 650. Each uses 300 paired world-model and simulator trials with base seed 100000, up to 300 environment steps, and eight executed controls per policy prediction. For policy $j$, let $\bar q_j^{\mathrm{model}}$ and $\bar q_j^{\mathrm{sim}}$ be its trial-mean IoUs. The reported error is
\begin{equation}
\mathrm{MAE}_{\mathrm{policy}}=\frac17\sum_{j=1}^7
\left|\bar q_j^{\mathrm{model}}-\bar q_j^{\mathrm{sim}}\right|.
\end{equation}
Pearson correlation is computed over the seven pairs of means, not over all individual trials. Within each model evaluation, paired trials use corresponding initializations. Random-number consumption differs across samplers, so simulator score arrays are not asserted to be identical across all models.

Table~\ref{tab:new-policy} gives the policy-level means for released GPC and AVDC-derived, recomputed from their paired trial arrays. Correlation is descriptive because it summarizes only seven policy checkpoints.
\input{tables/new_policy}

The released-GPC adapter was checked against the official sampler across consecutive chunks, including the past-control history. The AVDC-derived baseline is the locally trained action-conditioned adaptation, not a released AVDC policy simulator. Its final checkpoint contains 9,840 updates. 

GPC-RANK uses a fixed proposal distribution, frozen outcome readers, and 50 environment seeds. Model predictions rank the action proposals. The selected controls are then executed in the simulator. The paired planning interval in the main text resamples environment seeds. It does not quantify variation from retraining the world models.

\subsection{Checkpoint Choice and Policy Calibration}\label{app:policy-stages}
The final spatial checkpoint is selected for the joint visual-prediction recipe, not by taking the best policy score from each stage. Earlier policy results remain useful evidence (Table~\ref{tab:policy-stages}). The spatial-five checkpoint has error $0.0098$ and $r=0.9462$, slightly better on these policy metrics than the final spatial-thirty model, although its PushT LPIPS is worse ($0.0407$ versus $0.0280$). The learned-direction branch has the highest correlation among these entries but much larger absolute score error. Correlation can preserve policy ordering despite biased scores. 

\begin{table}[ht]\centering\small\setlength{\tabcolsep}{7pt}
\caption{\textbf{Separate PushT policy and visual checkpoints.} Seven frozen policies and 300 paired trials each. LPIPS uses AR64. Entries are descriptive stage results, not independently selected best metrics of one model.}\label{tab:policy-stages}
\begin{tabular}{@{}lrrr@{}}\toprule
Checkpoint&Policy error $\downarrow$&Pearson $r$ $\uparrow$&LPIPS $\downarrow$\\\midrule
Spatial, 5 epochs&.0098&.9462&.0407\\
Spatial, 30 epochs (main)&.0100&.9280&.0280\\
Learned-direction alternative&.0355&.9743&.0501\\
\bottomrule\end{tabular}\end{table}

\subsection{Released-Model Prediction References}\label{app:released}
The released checkpoints were trained under different data and optimization settings and may have seen episodes in our evaluation split. We report them separately to make the comparison with stronger released models visible without claiming equal held-out exposure. The released Can DriftWorld checkpoint also uses different renders, which confounds its static appearance error.
\begin{table}[ht]\centering\small\setlength{\tabcolsep}{7pt}
\caption{\textbf{Released references on the RGB evaluator.} All errors are lower-is-better. These are not equal-training-data comparisons.}\label{tab:released}
\begin{tabular}{@{}llrrr@{}}\toprule
Task&Checkpoint&MSE&Moving MSE&LPIPS\\\midrule
PushT&Released DriftWorld&.0085&.1182&.0163\\
&Released GPC&.0192&.2446&.0324\\
&Final BWM&.0098&.1247&.0280\\\midrule
Can&Released DriftWorld&.0525&.0662&.0676\\
&Final BWM&.0022&.0242&.0056\\
\bottomrule\end{tabular}\end{table}

\subsection{Qualitative Comparisons}\label{app:qualitative}
\begin{figure}[ht]
\centering\includegraphics[width=\linewidth]{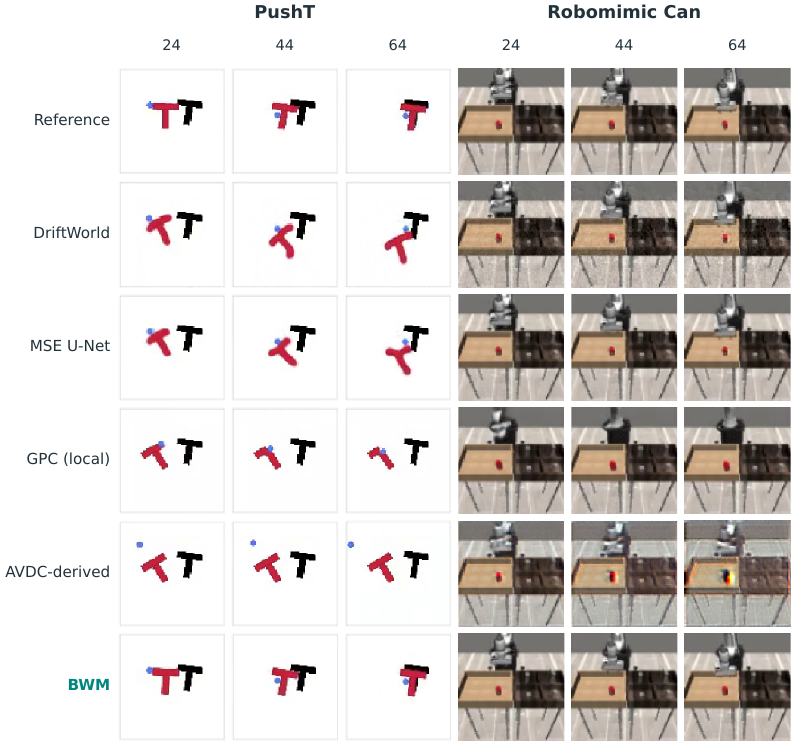}
\caption{\textbf{Additional RGB baseline comparisons.} The same two evaluation sequences as Figure~\ref{fig:rgb}, with local GPC and AVDC-derived rows included. BWM uses its final spatial continuation.}\label{fig:all-baselines}
\end{figure}

The RGB figures show two evaluation sequences: PushT saved sample 2 and Can saved sample 3, at frame indices 23, 43, and 63 (zero-based). They are illustrative selected sequences, not random samples of the evaluation distribution. The same frame indices and actions are used for all rows. Source-noise draws may differ across models. Figures use the displayed evaluation frames. Quantitative metrics are computed independently on the full episode cohorts. Figure~\ref{fig:all-baselines} includes the local GPC and AVDC-derived predictions omitted from the more compact main-text comparison.

%% file: tables/rgb_complete.tex
\begin{table}[ht]\centering\small\setlength{\tabcolsep}{4pt}
\caption{\textbf{Final BWM prediction metrics and episode-bootstrap intervals.} Each cell is a mean with its 95\% interval. The spatial stage has 30 epochs.}\label{tab:rgb-complete}
\begin{tabular}{@{}llcc@{}}\toprule
Protocol&Metric&PushT&Robomimic Can\\\midrule
64 frames & MSE $\downarrow$ & 0.0098 [0.0086, 0.0111] & 0.0022 [0.0016, 0.0028]\\
 & Moving MSE $\downarrow$ & 0.1247 [0.1098, 0.1401] & 0.0242 [0.0177, 0.0319]\\
 & LPIPS $\downarrow$ & 0.0280 [0.0242, 0.0320] & 0.0056 [0.0045, 0.0068]\\
 & SSIM $\uparrow$ & 0.9754 [0.9731, 0.9776] & 0.9843 [0.9812, 0.9871]\\
 & PSNR $\uparrow$ & 31.21 [30.82, 31.61] & 34.75 [33.79, 35.69]\\
\midrule
Full episode & MSE $\downarrow$ & 0.0154 [0.0137, 0.0171] & 0.0082 [0.0057, 0.0112]\\
 & Moving MSE $\downarrow$ & 0.1646 [0.1475, 0.1829] & 0.0517 [0.0379, 0.0680]\\
 & LPIPS $\downarrow$ & 0.0412 [0.0362, 0.0465] & 0.0238 [0.0160, 0.0338]\\
 & SSIM $\uparrow$ & 0.9662 [0.9634, 0.9689] & 0.9560 [0.9442, 0.9658]\\
 & PSNR $\uparrow$ & 28.67 [28.26, 29.08] & 30.17 [29.00, 31.34]\\
\bottomrule\end{tabular}\end{table}

%% file: tables/new_policy.tex
\begin{table}[ht]\centering\small\setlength{\tabcolsep}{8pt}
\caption{\textbf{PushT baseline policy estimates.} World-model and simulator mean IoUs for each frozen policy. Each entry averages 300 trials.}\label{tab:new-policy}
\begin{tabular}{@{}rrrrr@{}}\toprule
&\multicolumn{2}{c}{Released GPC}&\multicolumn{2}{c}{AVDC-derived}\\
\cmidrule(lr){2-3}\cmidrule(lr){4-5}
Policy epoch&Model&Simulator&Model&Simulator\\\midrule
50 & 0.5105 & 0.6190 & 0.1800 & 0.6103\\
100 & 0.5100 & 0.5929 & 0.1770 & 0.5922\\
150 & 0.4939 & 0.6029 & 0.1829 & 0.6042\\
200 & 0.4862 & 0.5630 & 0.1768 & 0.5609\\
250 & 0.4798 & 0.5536 & 0.1857 & 0.5507\\
300 & 0.4723 & 0.5617 & 0.1758 & 0.5624\\
650 & 0.4809 & 0.5423 & 0.1813 & 0.5448\\
\bottomrule\end{tabular}\end{table}

%% file: appendices/native.tex
\clearpage
\section{Native Video Implementation and Evaluation}\label{app:native}
\subsection{Representation and Losses}
The native backbone is DriftWorld's action-conditioned U-Net (175.4M parameters for Bridge-V2 and 160.3M for RT-1). Each call predicts one $16\times32\times32$ SD3 latent frame from a four-frame latent history and the corresponding action history. The VAE is frozen. The terminal map uses Gaussian source noise with standard deviation 1.35 and the same JVP-based stop-gradient target and stabilization as the RGB model. Source MSE operates on latents. Target-defined RGB motion masks are pooled to the latent grid. The range penalty acts on decoded RGB. Spatial VGG supervision compares decoded predictions with decoded target latents. There is no within-call temporal difference for a single predicted frame.

Recovery uses a valid eight-frame training sequence and one to three generated prefix frames. Prefixes are detached and appended to the history before predicting the next target. The recovery branch occurs every fourth update, with the same pixel and latent endpoint weighting as the parent. Feature drifting at recovery endpoints uses the preceding ground-truth frame to compute motion weights, so an erroneous generated history cannot redefine the target motion mask.

\paragraph{Feature drifting.}
For each condition, $K=8$ independent one-call samples are contrasted with one target using DriftWorld's retained drifting objective. The feature spaces are blocks 2, 5, and 8 of DINOv3 ViT-B with 16-pixel patches (a $16\times16$ spatial grid of 768-dimensional features) and the VAE latent grid (a $32\times32$ grid of 16-dimensional features). DINO and VAE parameters remain frozen, with gradients propagated to generated inputs. We retain the feature normalization, field normalization, multi-temperature aggregation, and motion weighting of the native implementation. The drifting coefficient is 0.01, both on clean histories and, in the refined arm, on recovery histories. Action accentuation is disabled and the inference guidance input is one.

The clean-history and recovery drifting terms each use new $K$-sample draws. The recovery term reuses the already-generated detached prefix, adds no second prefix rollout, and ramps from zero to 0.01 over 100 child updates. It is additional to the existing recovery reconstruction term. Activation checkpointing in the frozen VAE decoder reduces training memory. Neither feature encoder is required at inference, although the VAE decoder is needed to display latent predictions.

\subsection{Native Terminal-Map Objective}
For $256\times256$ RT-1 and Bridge-V2 videos, we predict one SD3 VAE latent frame per call \citep{esser2024sd3}, using the native action-conditioned DriftWorld backbone. Transport and source losses operate in latent space. Motion masks are pooled from ground-truth RGB differences, and spatial supervision compares decoded predictions with decoded target latents. Because each call predicts one frame, the within-chunk temporal loss is zero in this extension.

The richer appearance of these datasets motivates additional perceptual supervision. We use DriftWorld's conditional drifting loss in frozen DINOv3 and latent feature spaces \citep{simeoni2025dinov3,lu2026driftworld}, with $K=8$ generated samples per condition and coefficient $0.01$. The final refinement applies the same feature loss at existing generated-history recovery endpoints:
\begin{equation}
\mathcal L_{\mathrm{native}}=\mathcal L_{\mathrm{map+endpoint}}
+0.01\mathcal D(h,a,y)+0.01q_j I_k\mathcal D(\sg(\widehat h),a_{\mathrm{rec}},y_{\mathrm{rec}}),\label{eq:native}
\end{equation}
where $\mathcal D$ is the retained conditional feature-drifting loss and $q_j=\min((j+1)/100,1)$. The recovery feature term is added only when recovery is active and is not multiplied by the RGB recovery coefficient $0.25$. This is a hybrid training objective: it preserves the terminal map and one-call deployment, but includes a generated population during training. The RGB population-free cost measurements therefore do not describe this extension.

\subsection{Training and Controlled Continuations}
The native model first trains for 30 epochs on a common subset (2,078 Bridge-V2 episodes and 2,041 RT-1 episodes), then continues spatial training on full data through epoch 60. The full-data caches contain 52,969 Bridge-V2 and 86,634 RT-1 training episodes. Feature drifting is added through epoch 70. The selected recovery refinements start from these epoch-70 EMA weights and add 500 Bridge-V2 or 1,000 RT-1 updates. Their controls use the same parents and added update counts with recovery-feature weight zero.

Both arms use fresh Muon and AdamW optimizer states, a 100-update warmup, batch four per GPU, EMA decay 0.999, gradient clipping at two, and learning rates scaled to one quarter of the original learning rates of $10^{-3}$ and $10^{-4}$. The existing source, motion, recovery, and spatial ramps have completed before the refinement. The new recovery-feature coefficient is the changed loss term. The same seed and data setup do not imply identical later random draws because the added population consumes randomness. Added feature computation also makes these equal-update rather than equal-compute controls.

Bridge-V2 development compared several refinement checkpoints and coefficients. The 500-update full-weight result was selected after validation review. The RT-1 transfer selected the 1,000-update checkpoint. Both use one training seed. These results support the chosen continuation on the observed validation cohorts. They are not a preregistered, untouched-test estimate of its general effect.

\subsection{Native Evaluation Protocol}
The evaluation cohorts contain 215 Bridge-V2 episodes with 7,079 predicted frames and 198 RT-1 episodes with 8,409 predicted frames. Each eight-frame autoregressive window starts from four ground-truth latent frames. Longer displayed videos concatenate such windows. Reference images are obtained by decoding the target latents, and pixel and perceptual comparisons therefore include the VAE representation in both arms.

SSIM, PSNR, and LPIPS are averaged per episode and then equally across episodes. FID \citep{heusel2017fid} aggregates all generated and reference frames. FVD \citep{unterthiner2019fvd} uses the StyleGAN-V \texttt{fvd2048\_16f} evaluator on the first 16 generated frames of eligible episodes, with the same mirrored evaluation convention across arms. There are 201 eligible Bridge-V2 and 193 eligible RT-1 videos. FID and FVD values are retained evaluator outputs. They were not recomputed from compressed display videos. Tables~\ref{tab:native} and \ref{tab:native-complete} share the same episode cohorts within each dataset.
\input{tables/native_complete}

\paragraph{Paired perceptual differences.}
Aligning the control and selected-refinement records by episode identity gives 198 RT-1 and 215 Bridge-V2 pairs. We resample these pairs 10,000 times (seed 20261014), using each episode's mean LPIPS with equal weight. Recovery minus control LPIPS is $-0.03217$ on RT-1 (95\% percentile interval $[-0.03497,-0.02952]$), and $-0.02652$ on Bridge-V2 (95\% percentile interval $[-0.02816,-0.02500]$). Both descriptive intervals are below zero. They measure variation within the observed validation cohorts, not uncertainty from training seeds or checkpoint selection. The selected refinement remains without its own untouched-test confirmation. FID and FVD are distributional evaluator outputs, so we do not obtain intervals by resampling their aggregate scores.

\subsection{Common-Subset Baselines}\label{app:native-subset}
Table~\ref{tab:native-subset} retains the separate 30-epoch subset comparison. All methods see the same eligible training episodes and one deterministic crop per episode per epoch, use SD3 latents and the same native generator, and evaluate on the cohorts above. Their objectives, optimizer recipes, and batch sizes differ. These rows cannot be compared as equal-compute runs or pooled with the full-data continuations.

The BWM spatial model attains higher PSNR than DriftWorld on both tasks, but DriftWorld has better LPIPS, FID, and FVD. MSE training is also competitive in PSNR. This experiment demonstrates the trade-off of paired endpoint supervision and motivates the later feature-drifting extension. It does not support a claim that the BWM spatial recipe alone dominates population-based perceptual learning at native resolution.
\input{tables/native_subset}

\subsection{Selected Recovery Video Frames}\label{app:native-qualitative}
\begin{figure}[t]
\centering\includegraphics[width=\linewidth]{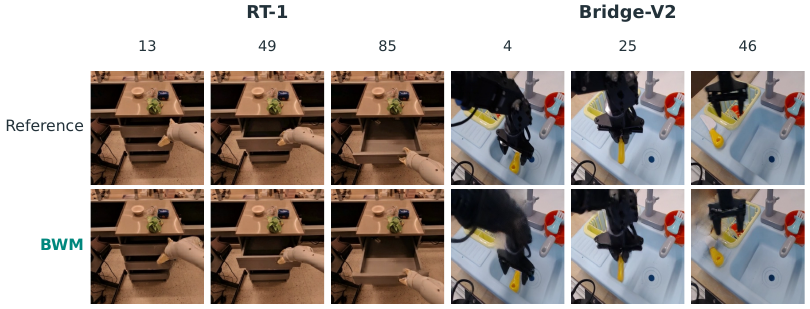}
\caption{\textbf{Real robotics video prediction.} RT-1 drawer manipulation and Bridge-V2 gripper motion from the final recovery-refined checkpoints. Columns show separated positions in each displayed video (positions 13, 49, and 85 for RT-1 and positions 4, 25, and 46 for Bridge-V2). Predictions are generated within the conditioned windows described in Section~\ref{sec:native}. References are VAE reconstructions.}
\label{fig:native-original}
\end{figure}

Figure~\ref{fig:native-original} retains the original selected recovery display. Figure~\ref{fig:native} augments the same episode positions with matched control predictions.

\subsection{Separate Paired-Feature Recovery and Independent Test}\label{app:paired-native}
A subsequent fidelity experiment replaces the \emph{added generated-history population term} with paired frozen-feature appearance and displacement supervision. The original clean-history $K=8$ drifting loss remains active. This is a different recipe from the selected recovery-feature refinements in Table~\ref{tab:native}. Its results cannot be pooled with those gains.

Bridge-V2 selects the 1,000-update EMA using validation gates before testing. The untouched test cohort has 213 episodes and 7,044 predicted frames, with the same eight-frame conditioned windows, evaluation seed 314, VAE-decoded references, and detector conventions for parent and child. Evaluation settings and reference frames match across arms. Table~\ref{tab:paired-native} retains both validation and independent-test outcomes. The test LPIPS improvement is $4.5\%$, while FVD worsens by $7.8\%$. FID improves $2.8\%$ and PSNR decreases slightly. The RT-1 candidate fails the 400-update validation gate and is not promoted to a test evaluation.

\begin{table}[ht]\centering\small\setlength{\tabcolsep}{6pt}
\caption{\textbf{Paired-feature variant: validation and independent test.} Bridge-V2 parent versus the separately selected 1,000-update EMA. This is not the 500-update recovery-population variant in Table~\ref{tab:native}. FID and FVD are retained evaluator outputs.}\label{tab:paired-native}
\begin{tabular}{@{}llrrrrr@{}}\toprule
Split&Model&PSNR $\uparrow$&SSIM $\uparrow$&LPIPS $\downarrow$&FID $\downarrow$&FVD $\downarrow$\\\midrule
Validation&Parent&25.00&.8555&.11125&31.86&175.53\\
&Paired-feature child&24.95&.8577&.10642&31.36&188.04\\\midrule
Test&Parent&25.53&.8631&.10439&32.69&160.06\\
&Paired-feature child&25.47&.8651&.09964&31.77&172.55\\
\bottomrule\end{tabular}\end{table}

All 16 frames in each of four fixed diagnostic windows were reviewed at the retained Bridge checkpoints. Changes are small, with late-block blur around the gripper and manipulated object and reset recovery persisting. No convincing flicker fix is established. This mixed result sharpens the interpretation of the main refinement: paired appearance improvements alone do not ensure improved video-level distributional quality. It does not identify the population term as the sole cause, because the variant also changes recovery features and selection history. The larger selected LPIPS, FID, and FVD gains still need their own untouched-test confirmation.

%% file: tables/native_complete.tex
\begin{table}[ht]\centering\small\setlength{\tabcolsep}{5pt}
\caption{\textbf{Complete native continuation metrics.} Same cohorts and models as Table~\ref{tab:native}.}\label{tab:native-complete}
\begin{tabular}{@{}llrrrrr@{}}\toprule
Dataset&Arm&SSIM $\uparrow$&PSNR $\uparrow$&LPIPS $\downarrow$&FID $\downarrow$&FVD $\downarrow$\\\midrule
RT-1 & Parent & 0.8840 & 27.35 & 0.0917 & 19.77 & 167.50\\
 & Control & 0.8836 & 27.34 & 0.0933 & 20.56 & 172.89\\
 & Recovery & 0.8861 & 27.27 & 0.0612 & 8.05 & 104.91\\
\midrule
Bridge-V2 & Parent & 0.8555 & 25.00 & 0.1113 & 31.86 & 175.53\\
 & Control & 0.8556 & 25.01 & 0.1106 & 30.98 & 171.56\\
 & Recovery & 0.8580 & 24.95 & 0.0841 & 21.34 & 117.20\\
\bottomrule\end{tabular}\end{table}

%% file: tables/native_subset.tex
\begin{table}[ht]\centering\small\setlength{\tabcolsep}{4.5pt}
\caption{\textbf{Thirty-epoch common-subset native comparison.} The same training episode subsets and evaluation cohorts are used within each dataset. Batches and optimizer recipes differ. These models precede BWM's full-data and recovery continuations.}\label{tab:native-subset}
\begin{tabular}{@{}llrrrrr@{}}\toprule
Dataset&Method&Calls per frame&PSNR $\uparrow$&LPIPS $\downarrow$&FID $\downarrow$&FVD $\downarrow$\\\midrule
RT-1 & DriftWorld & 1 & 22.42 & 0.1295 & 27.48 & 237.45\\
 & MSE U-Net & 1 & 23.03 & 0.1788 & 62.41 & 419.16\\
 & GPC-derived & 3 & 20.69 & 0.2937 & 124.18 & 884.52\\
 & AVDC-derived & 100 & 20.27 & 0.1751 & 94.87 & 719.10\\
 & BWM spatial & 1 & 23.15 & 0.1659 & 75.53 & 535.59\\
\midrule
Bridge-V2 & DriftWorld & 1 & 21.92 & 0.1042 & 14.18 & 188.13\\
 & MSE U-Net & 1 & 22.44 & 0.1663 & 61.96 & 405.61\\
 & GPC-derived & 3 & 19.93 & 0.3350 & 132.93 & 756.50\\
 & AVDC-derived & 100 & 19.59 & 0.1590 & 79.44 & 657.46\\
 & BWM spatial & 1 & 22.13 & 0.1550 & 68.86 & 592.24\\
\bottomrule\end{tabular}\end{table}

%% file: appendices/diagnostics.tex
\clearpage
\section{Terminal-Map and Endpoint Diagnostics}\label{app:diagnostics}
\subsection{Scope of the Terminal-Map Formulation}
The BTM transport theorem selects a terminal map through the autonomous flow and its terminal behavior \citep{lee2026btm}. Merely imposing a pointwise identity on the target and conservation away from it does not prove uniqueness without an appropriate terminal trace condition. Likewise, the population consistency argument for the unweighted stopped update is not a convergence theorem for our adaptively weighted, source-regularized neural objective. Our claims about the learned map are empirical.

\subsection{Fixed-Window Source Diagnostic}
The Can probe evaluates 90 fixed evaluation windows and four shared source draws per window at $s\in\{0,.25,.5,.75,1\}$. All conditions and targets are fixed across the compared models. Table~\ref{tab:source-diagnostic} gives the retained endpoint MSEs. Source anchoring primarily improves prediction from source noise, reducing its MSE by $70.9\%$ from the stabilized model. The continuation with motion supervision and generated-history recovery reduces it by $82.2\%$ from that same baseline. These checkpoints precede the final spatial continuation.
\begin{table}[ht]\centering\small\setlength{\tabcolsep}{6pt}
\caption{\textbf{Endpoint MSE across input locations on Can.} The source is $s=0$ and the target boundary is $s=1$.}\label{tab:source-diagnostic}
\begin{tabular}{@{}lrrrrr@{}}\toprule
Model&0&.25&.50&.75&1\\\midrule
Stabilized BTM&.002831&.002027&.001437&.000810&.000105\\
$+$ source&.000823&.001429&.001178&.000687&.000096\\
$+$ motion and recovery&.000503&.000868&.000747&.000498&.000025\\
Learned direction&.001170&.002241&.002476&.002513&.000112\\
\bottomrule\end{tabular}\end{table}

% \subsection{Effect of Source Supervision}
% To obtain Equation~\ref{eq:variance}, write $P=\mu_P+(P-\mu_P)$ and $y=\mu_y+(y-\mu_y)$ conditional on $c$. Expanding the squared norm gives the squared mean difference, the two conditional variances, and cross terms. The centered terms have zero conditional mean; conditional independence of $z$ and $y$ makes the covariance term vanish. Dividing by $D$ yields the stated decomposition. Consequently, source regression encourages a concentrated predictor around the conditional mean, even if the underlying data distribution is multimodal. It should not be interpreted as a proper distributional scoring rule for the full conditional law.

% The retained source-draw and action-swap probe evaluates eight noise draws on 64 fixed windows per task. The earlier five-epoch spatial checkpoint has $20.9\times$ less moving-region output spread than local DriftWorld on PushT, while its action-swap penalty remains large ($0.6447$ versus $0.5339$). These results show that low source variation can coexist with action sensitivity. They describe that earlier checkpoint; we do not relabel them as measurements of the final 30-epoch model.

\subsection{Endpoint Supervision}
Figure~\ref{fig:training-signals} gives the four complementary signals used around the deployed endpoint: changed pixels, temporal differences, frozen spatial features, and recovery under generated histories. Its image tiles illustrate the loss construction using real evaluation crops. They are not a captured training branch or an isolated loss ablation.

\subsection{Conditional Variance of the Source Loss}
This loss gives direct supervision to the model's one-call prediction. For source noise independent of the ground-truth future frames conditional on $c$, let $P=T_\theta(z,c)$. Then
\begin{equation}
\E[\normD{P-y}\mid c]=
\normD{\E[P\mid c]-\E[y\mid c]}+\VarD(P\mid c)+\VarD(y\mid c).\label{eq:variance}
\end{equation}
Here $\VarD$ denotes the mean conditional variance per coordinate. Source anchoring consequently penalizes both conditional-mean error and variation induced by source noise. This is useful for paired prediction in largely deterministic manipulation settings, but can suppress valid conditional diversity. We evaluate prediction accuracy rather than claim calibrated multimodal coverage.

% \subsection{Cumulative Recipe Progression}\label{app:progression}
% \input{tables/ablation}
% \paragraph{Source anchoring.}
% The fixed-window Can probe holds the history and actions fixed while changing the transport input. Source MSE decreases from $0.002831$ for the stabilized map to $0.000823$ after source anchoring, and to $0.000503$ after motion/recovery. Boundary error is already small ($0.000105$ before anchoring). Thus the important change occurs at the source input used during inference. Table~\ref{tab:ablation} shows corresponding improvements in autoregressive LPIPS.

% \paragraph{Endpoint regularization.}
% Adding motion and recovery supervision reduces full-episode MSE from $0.0940$ to $0.0112$ on Can and from $0.0434$ to $0.0299$ on PushT. Spatial supervision improves short-horizon LPIPS further, and continued spatial training yields the final $0.0280$ and $0.0056$. These stages are cumulative: the table records their observed progression, rather than attributing the entire difference to an isolated loss at fixed compute.

% \paragraph{Learned-direction alternative.}
% An alternative branch fits a direction model from the source-anchored parent. It reaches PushT LPIPS $0.0501$ but Can LPIPS $0.0854$, worse than the simpler source-anchored Can map. We therefore retain the direct sampled pair direction. This comparison supports the simpler recipe in our tasks, without implying that learned directions are generally inferior.

%% file: appendices/fresh.tex
\clearpage
\section{Fresh RGB Training and Failure Analysis}\label{app:fresh}
\subsection{Matched Clean-Window Comparison}
To check how much of the final comparison depends on inherited training, we also train a fresh BWM generator for the same five capped epochs and clean-window exposure as the audited local MSE and DriftWorld fits. All warmup fits inside the budget. The predeclared final EMA is evaluated without checkpoint selection. Recovery adds supervision and the optimizer differs, so this controls clean exposure rather than total computation. Table~\ref{tab:fresh} reports the 64-frame results.
\input{tables/fresh}
Relative to MSE, fresh BWM lowers PushT pixel and moving-region MSE by $30\%$ and $29\%$, but has $22\%$ worse LPIPS. Can pixel MSE improves by $6\%$ over 60 predicted frames, whereas full-episode MSE is $14\%$ worse and LPIPS is $104\%$ worse.  The controlled result supports useful PushT prediction at limited clean exposure, while showing that one-call training alone does not guarantee sharp, sustained dynamics.

\subsection{Budget and Selection}
This separate experiment initializes the generator, EMA, and optimizer from scratch at seed 1, with zero inherited updates. Frozen pretrained VGG16 remains an auxiliary training loss. It is not a pretrained generator. Can completes 20,000 updates at global batch two (40,000 clean windows, 10,000 additional recovery targets). PushT completes 39,350 updates at global batch four (157,400 clean windows, 39,348 additional recovery targets). These are five \emph{capped} epochs, corresponding to 0.719 and 0.409 full-corpus clean pass equivalents, respectively. 

The sampler, split, backbone, four-frame history, action alignment, and clean-window exposure match the audited local five-epoch baselines. Existing baseline final checkpoints are reevaluated with the same numerical protocol. The two-rank seed-1 BWM runs use Muon at $0.00025$ plus AdamW at $0.000025$, whereas the baselines use AdamW ($0.000025$ on Can, $0.0000125$ on PushT). Source weight is one from the first update. Optimizer warmup takes 500 updates, and recovery, prefix-depth, and spatial ramps take 1,000, all within the stated budget. No held-out tuning, early stopping, or best-checkpoint promotion follows these runs: the final epoch-five EMA is predeclared. Thus the comparison matches clean exposure, with additional BWM endpoint supervision and different optimization. Historical baseline per-update timing and exact training-code revisions were not retained, so this experiment supplies no controlled training-speed estimate.

%% file: tables/fresh.tex
\begin{table}[H]\centering\small\setlength{\tabcolsep}{6pt}
\caption{\textbf{Fresh generators at matched clean-window exposure.} 64-frame evaluation after five capped epochs. BWM has zero inherited updates. Recovery targets, optimizer, and compute differ. The predeclared final checkpoint is used, separately from Table~\ref{tab:rgb}.}\label{tab:fresh}
\begin{tabular}{@{}llrrr@{}}\toprule
Task&Model&MSE $\downarrow$&Moving MSE $\downarrow$&LPIPS $\downarrow$\\\midrule
PushT&Local DriftWorld&.0314&.4093&\textbf{.0790}\\
&MSE U-Net&.0303&.4008&.0804\\
&Fresh BWM&\textbf{.0211}&\textbf{.2832}&.0980\\
\midrule
Can&Local DriftWorld&.0104&.0769&.0682\\
&MSE U-Net&.0051&.0544&\textbf{.0104}\\
&Fresh BWM&\textbf{.0048}&\textbf{.0475}&.0154\\
\bottomrule\end{tabular}\end{table}